\documentclass[journal]{IEEEtran}
\usepackage{amsmath,amsfonts}
\usepackage{algorithmic}
\usepackage{algorithm}
\usepackage{array}
\usepackage[caption=false,font=normalsize,labelfont=sf,textfont=sf]{subfig}
\usepackage{textcomp}
\usepackage{stfloats}
\usepackage{url}
\usepackage{verbatim}
\usepackage{graphicx}
\usepackage{xurl}
\usepackage{cite}
\newcommand\IEEEpreprint{
    \let\thefootnote\relax\footnotetext{This work has been submitted to the IEEE for possible publication. Copyright may be transferred without notice, after which this version may no longer be accessible.}
}
\begin{document}

\title{Mechanistic Decoding of Cognitive Constructs in Large Language Models}
\author{Yitong~Shou,~Manhao~Guan%
\thanks{The authors are with the College of Computer Science and Technology, Zhejiang University, Hangzhou 310027, China.}%
\thanks{Corresponding author: Manhao Guan (e-mail: guanmh@zju.edu.cn).}
}
\markboth{Preprint submitted to IEEE}%
{Shou and Guan: Mechanistic Decoding of Cognitive Constructs in Large Language Models}
\maketitle
\IEEEpreprint
\begin{abstract}
While Large Language Models (LLMs) demonstrate increasingly sophisticated affective capabilities, the internal mechanisms by which they process complex emotions remain unclear.
Existing interpretability approaches often treat models as black boxes or focus on coarse-grained basic emotions, leaving the cognitive structure of more complex affective states underexplored.
To bridge this gap, we propose a Cognitive Reverse-Engineering framework based on Representation Engineering (RepE) to analyze social-comparison jealousy. 
By combining appraisal theory with subspace orthogonalization, regression-based weighting, and bidirectional causal steering, we isolate and quantify two psychological antecedents of jealousy, \textit{Superiority of Comparison Person} and \textit{Domain Self-Definitional Relevance}, and examine their causal effects on model judgments. 
Experiments on eight LLMs from the Llama, Qwen, and Gemma families suggest that models natively encode jealousy as a structured linear combination of these constituent factors. 
Their internal representations are broadly consistent with the human psychological construct, treating \textit{Superiority} as the foundational trigger and \textit{Relevance} as the ultimate intensity multiplier. 
Our framework also demonstrates that toxic emotional states can be mechanically detected and surgically suppressed, suggesting a possible route toward representational monitoring and intervention for AI safety in multi-agent environments.
\end{abstract}


\section{Introduction}
\label{sec:introduction}
The advent of Large Language Models (LLMs) has catalyzed a paradigm shift in Affective Computing (AC). Moving beyond simple sentiment classification, modern LLMs show strong capabilities in affective understanding and generation, supporting applications such as mental health support, companion AI, and interactive non-player characters (NPCs) \cite{zhangAffectiveComputingEra2026a}. As these models are increasingly deployed in socially sensitive contexts, it is important to ensure that their emotional reasoning is transparent, predictable, and aligned with human values.

Despite these advancements, the internal mechanisms by which LLMs process emotions remain unclear. Inference-time methods such as prompt engineering and chain-of-thought prompting rely on textual outputs, but such outputs may constitute unfaithful post-hoc rationalizations. Parameter-adaptation methods such as LoRA fine-tuning improve downstream performance, yet they still treat the model's internal representations as an optimization target rather than an object of cognitive explanation.

Interpretability research offers a possible route forward, but existing paradigms still leave a granularity gap. Traditional bottom-up approaches, such as circuit analysis \cite{anthropic2025biology} and sparse autoencoders \cite{templeton2024scaling}, operate at the level of neurons or local features and are often too microscopic for abstract psychological constructs. Representation Engineering (RepE), by contrast, isolates high-level concepts that emerge from population-level neural activity \cite{zou2023representation}. Although Tak et al. \cite{takMechanisticInterpretabilityEmotion} showed that RepE can probe coarse-grained basic emotions, it remains unclear whether the same paradigm can decode complex emotions whose elicitation depends on specific psychological antecedents rather than broad appraisal dimensions.

This challenge is especially salient for \textbf{social-comparison jealousy}. Rather than reflecting general positive or negative valence, jealousy arises from upward social comparison. A mechanistic account therefore requires disentangling the specific antecedents that produce this emotion, not merely relying on general appraisal dimensions. In this study, we focus on two theoretically central antecedents---\textit{Superiority of Comparison Person} and \textit{Domain Self-Definitional Relevance}---and include \textit{Weekday} as a placebo control to test whether the extracted representations reflect meaningful psychological structure.

To address this challenge, we propose a \textbf{Cognitive Reverse-Engineering} framework that extends RepE to support fine-grained affective analysis. The framework comprises four phases:
\begin{itemize}
    \item \textit{Phase I: Representation Extraction and Validation}
    \item \textit{Phase II: Subspace Orthogonalization}
    \item \textit{Phase III: Statistical Weighting and Validity Check}
    \item \textit{Phase IV: Causal Intervention Framework}
\end{itemize}

We use this framework to answer three core research questions (RQs):

\begin{itemize}
    \item \textbf{RQ1:} What are the internal causal weights of jealousy's constituent factors? (Addressed via \textit{Phase III})

    \item \textbf{RQ2:} Where do representations of jealousy and its constituent factors emerge and stabilize within the network?  (Addressed via \textit{Phase I \& IV})

    \item \textbf{RQ3:} Can we directionally manipulate the model's jealousy judgments using the extracted factor representations? (Addressed via \textit{Phase IV})

\end{itemize}

Based on evaluations across three major model families (Llama, Qwen, and Gemma), the main contributions of this study are as follows:

\begin{enumerate}
    \item \textbf{Methodological Contribution to Interpretability:} Recognizing the limitations of bottom-up mechanistic approaches for abstract psychological phenomena, we adopt a top-down Representation Engineering (RepE) approach to capture population-level, high-dimensional representations. We further extend traditional coarse-grained RepE with the \textbf{Cognitive Reverse-Engineering} framework. This framework incorporates a combinatorial dataset strategy with placebo controls, subspace orthogonalization for feature purification, and a shift from qualitative steering to quantitative regression analysis and causal interventions.

    \item \textbf{First White-Box Semantic Dissection of Complex Emotion in LLMs:} While prior studies have evaluated jealous behaviors in large models externally \cite{huangApatheticEmpatheticEvaluating2024}, they have treated the models largely as black boxes. This study instead provides a white-box analysis of the internal cognitive logic behind social-comparison jealousy. 
    Grounded in human psychological theories and empirical research, we dissect this complex emotion into high-dimensional semantic factors---\textit{Superiority} and \textit{Relevance}. In Phase III (addressing RQ1), we show that LLMs represent this emotion as a structured linear combination of these factors, which aligns with human emotional logic: treating \textit{Superiority} as the initial trigger and \textit{Relevance} as the ultimate intensity multiplier.

    \item \textbf{Implications for AI Safety:} For RQ2, our layer-wise profiling (Phase I \& IV) identifies the mid-to-late layers as critical zones where representations become stable. This enables \textbf{real-time representational monitoring} to probe the AI's latent space for deceptive alignment. For RQ3, our causal intervention (Phase IV) successfully manipulates the model behaviors via bidirectional steering, using factors' representation vectors. In multi-agent environments, surgically suppressing these vectors may help reduce harmful social-comparison behaviors and competitive dynamics \cite{panRewardsJustifyMeans2023}.

\end{enumerate}

\section{Psychological Background}
\label{sec:psychological background}

\subsection{Conceptualizing Social-Comparison Jealousy}

In psychological research, the experience of jealousy is not a monolithic construct. A critical distinction is drawn between \textit{social-relations jealousy} and \textit{social-comparison jealousy}. 

According to the seminal framework proposed by Salovey and Rodin \cite{saloveyAntecedentsConsequencesSocialComparison}, \textbf{social-relations jealousy} typically arises in romantic or interpersonal contexts, involving a ``desire for exclusivity'' and the suspicion that a desired relationship is threatened by a rival.

In contrast, this study focuses on \textbf{social-comparison jealousy}, which is defined as a state involving a ``desire for superiority on some dimension'' \cite{saloveyAntecedentsConsequencesSocialComparison}. This form of jealousy is triggered when an individual receives negative feedback about their own performance or attributes relative to a successful other. Although social-comparison jealousy is philosophically and semantically distinct from ``envy''---traditionally defined as discontent with one's own situation and a desire for another's attributes \cite{brysonModesResponseJealousyevoking1991}---Salovey and Rodin argue that laypeople often use the terms interchangeably. Furthermore, the emotional and cognitive reactions associated with being out-performed in a self-relevant domain are functionally similar whether labeled as envy or jealousy.

Therefore, following this precedent, we use the term \textit{\textbf{social-comparison jealousy}} to specifically denote the affective state resulting from an upward social comparison, where one feels diminished by another's superiority. In the remainder of this paper, the terms ``envy'' and ``jealousy'' are used interchangeably to refer to this specific social-comparison construct, distinct from romantic possessiveness.

\subsection{Theoretical Frameworks and Antecedents}
This study investigates the antecedents of social-comparison jealousy. We synthesized relevant theories---specifically Social Comparison Theory---along with empirical findings to comprehensively cover the factors that precipitate jealousy. 

Based on Social Comparison Theory \cite{festingerTheorySocialComparison1954,alickeSocialComparisonEnvy2008,corcoranSocialComparisonMotives2011} and the Self-Evaluation Maintenance (SEM) model \cite{tesserEmotionSocialComparison1991a}, an upward social comparison is the basic requirement of envy. We identified four key factors rooted in these theoretical frameworks.

\begin{itemize}
    \item \textbf{Superiority of Comparison Person\cite{takahashiWhenYourGain2009,smithRolesOutcomeSatisfaction1990,crusiusEnvyAdversarialReview2020}.} This refers to the presence of another person who enjoys a superior outcome or advantage in the \textit{same domain} where the individual has experienced a setback. Smith et al. emphasize that this superiority is most potent when the domain is self-relevant \cite{smithRolesOutcomeSatisfaction1990}.

    \item \textbf{Domain Self-Definitional Relevance\cite{saloveyAntecedentsConsequencesSocialComparison,tesserEmotionSocialComparison1991a,lockwoodSuperstarsMePredicting1997,saloveyProvokingJealousyEnvy1991,smithComprehendingEnvy2007,crusiusEnvyAdversarialReview2020}.} Not all failures trigger jealousy. According to Salovey and Rodin \cite{saloveyAntecedentsConsequencesSocialComparison}, a \textit{self-definitionally relevant domain} refers to a performance domain that is particularly ``self-involving'' or central to the individual's ``self-schema.'' Jealousy is most intense when the negative feedback occurs in a domain that the individual uses to define who they are.

    \item \textbf{Similarity with Comparison Person\cite{takahashiWhenYourGain2009,ComparingLotsPromotion2004,smithComprehendingEnvy2007,crusiusEnvyAdversarialReview2020}.} Envy is not directed at random targets. Smith and Kim \cite{smithComprehendingEnvy2007} argue that we envy people who are ``similar to ourselves, save for their advantage on the desired domain.'' This similarity often includes comparison-related attributes, making the comparison psychologically relevant.

    \item \textbf{Comparison Alternative\cite{smithRolesOutcomeSatisfaction1990}.} This factor examines the broader context of an individual's self-worth. It is defined as the level of outcomes a person experiences on the ``next most favourable dimension of comparison'' (e.g., another valued domain like social life or career). A favorable comparison alternative allows one to compensate for inferiority in the primary domain, whereas an unfavorable one leaves the individual with no means to buffer their self-esteem \cite{smithRolesOutcomeSatisfaction1990}.

\end{itemize}

Among the four distinct factors identified in the literature, a theoretical distinction must be drawn between the \textbf{primary driving mechanisms} and \textbf{secondary moderating variables}. We argue that Superiority of Comparison Person and Domain Self-Definitional Relevance constitute the two core mechanisms for the elicitation of social-comparison jealousy. 

This argument follows from the definition established earlier: social-comparison jealousy involves a ``desire for superiority'' that is triggered when one falls short in a specific, self-relevant context \cite{saloveyAntecedentsConsequencesSocialComparison,tesserEmotionSocialComparison1991a}. In this framework, \textit{Superiority} acts as the indispensable initiator (the trigger); without it, there is no threatening upward comparison. Once triggered, \textit{Relevance} serves as the primary amplifier; without it, an upward comparison is more likely to produce indifference or even basking in reflected glory than distress. 

In contrast, the remaining factors primarily act as secondary moderators that fine-tune the intensity of the emotion, rather than determining its emergence and core intensity.

Therefore, as a foundational investigation into the encoding of envy in Large Language Models, this study prioritizes the isolation and analysis of these two core mechanisms. By focusing on \textbf{Superiority} and \textbf{Relevance}, we aim to characterize the basic representational structure of social-comparison jealousy in LLMs before examining the more nuanced effects of other variables.

\section{Related Work}
\subsection{Current Landscape of Affective Computing in LLMs}

The advent of Large Language Models (LLMs) has catalyzed a paradigm shift in Affective Computing (AC), moving from fine-tuning small Pre-trained Language Models (PLMs) (e.g., BERT, RoBERTa) to leveraging the generative capabilities of LLMs for both Affective Understanding (AU) and Affective Generation (AG) tasks \cite{zhangAffectiveComputingEra2026a}. Current research paradigms can be broadly categorized into inference-time optimization (black-box) and parameter adaptation (white-box).

\subsubsection{Inference-Time Optimization (Black-box Approaches)}

To use the emotional reasoning capabilities of LLMs without accessing internal weights, researchers have widely adopted prompt engineering and in-context learning. Notable strategies include Chain-of-Thought (CoT) prompting, which guides models to generate intermediate reasoning steps for complex tasks such as Implicit Sentiment Analysis (e.g., THOR \cite{fei2023reasoning}) or Emotion Cause Extraction \cite{wu2024enhancing}. Furthermore, Agent-based frameworks (e.g., Agent4SC \cite{zhang2024agent4sc}) have been proposed to simulate multi-turn social interactions and emotional dynamics among multiple LLMs.

\textbf{Limitations:} However, these black-box methods are limited by the problem of interpretive ``unfaithfulness.'' Research suggests that the natural language reasoning generated by LLMs is often a post-hoc rationalization rather than a reflection of the model's true decision-making process \cite{turpin2024language}. Relying solely on output text fails to verify whether the model's internal logic aligns with established psychological theories of emotion.

\subsubsection{Parameter Adaptation (White-box Approaches)}

To enhance performance on specific affective domains, recent studies have adopted Instruction Tuning and Parameter-Efficient Fine-Tuning (PEFT) (e.g., LoRA). For instance, EmoLLM \cite{liu2024emollm} and InstructERC \cite{lei2024instructerc} fine-tune general-purpose LLMs on comprehensive emotional datasets, achieving state-of-the-art results in mental health counseling and emotion recognition in conversation. These approaches are technically ``white-box'' as they involve updating model parameters.

\textbf{Limitations:} Nevertheless, existing white-box research remains \textbf{performance-oriented} rather than \textbf{mechanism-oriented}. While fine-tuning optimizes surface-level metrics (e.g., F1-score), it still treats the model's internal representation as an opaque optimization target. It alters \textit{how} the model behaves but fails to elucidate \textit{why} the model predicts specific emotions from a cognitive perspective. This underscores the need for a deeper mechanistic approach.

\subsection{Interpretability Methods}
\label{sec:related_work_mechanistic}

To elucidate \textit{why} models predict specific emotions from a cognitive perspective, current interpretability methodologies can be broadly categorized into two paradigms: the bottom-up approach and the top-down approach.

\subsubsection{Bottom-Up: Mechanistic Interpretability (MI)}
Mechanistic Interpretability represents a \textbf{bottom-up} approach to transparency. Aligning with the \textit{Sherringtonian view} in cognitive neuroscience \cite{zou2023representation}, it focuses on the microscopic level, attempting to reverse-engineer neural networks into specific ``circuits'' or algorithms. 

Several classic methodologies form the foundation of this literature:

\begin{itemize}
    \item \textbf{Causal Intervention / Causal Tracing:} This technique involves perturbing parameters or hidden vectors at specific locations within the model and measuring the corresponding variations in the final output. The magnitude of these changes is used to localize the components most critical to the model's predictions \cite{meng2022locating, comparative_neuron_2024}.
    
    \item \textbf{Logit Lens and Neuron Analysis:} This approach leverages the observation that Feed-Forward Network (FFN) neurons exhibit strong interpretability when projected into the unembedding (vocabulary) space. By correlating these projected conceptual representations with the final output, researchers can pinpoint the exact neurons that promote specific predictions \cite{geva2022transformer, neuron_knowledge_2024}.
    
    \item \textbf{Circuit Analysis:} This method treats learned features as fundamental units to construct end-to-end computational circuits (or attribution graphs) from input to output, meticulously mapping the flow of information through the network's layers and attention heads \cite{anthropic2025biology, dunefsky2024transcoders}.

    \item \textbf{Sparse Autoencoders (SAE):} Building upon neuron analysis, researchers identified the phenomenon of \textit{superposition}—where models pack many unrelated concepts into fewer neurons, resulting in polysemanticity \cite{elhage2022toy}. To untangle these representations, SAEs are trained to map the dense activations of LLMs into a higher-dimensional, sparse dictionary space, thereby extracting highly abstract and monosemantic features \cite{templeton2024scaling}.
\end{itemize}

\subsubsection{Top-Down: Representation Engineering (RepE)}
In contrast to microscopic circuit discovery, Representation Engineering (RepE) advocates for a \textbf{top-down} approach. Drawing on the \textit{Hopfieldian view} \cite{zou2023representation}, it posits that cognition is an emergent product of representational spaces implemented by population-level neural activity. RepE places \textbf{latent representations} at the center of analysis, allowing researchers to monitor and causally manipulate high-level cognitive phenomena by analyzing and steering hidden states.

\subsection{Existing Gaps and Methodological Rationale}
\label{sec:existing_gaps}

Prior work has applied both interpretability paradigms to emotion inference and generation. For instance, Tak et al. \cite{takMechanisticInterpretabilityEmotion} utilized the top-down RepE paradigm to probe and steer general appraisal concepts, while recent work by Wang et al. \cite{wangLLMsFeelEmotion2025} leveraged bottom-up approaches to discover and control microscopic ``emotion circuits'' (specific MLP neurons and attention heads). Together, these studies suggest that LLMs construct structured internal representations for \textbf{coarse-grained basic emotions} (e.g., joy, sadness, and anger).

While these studies established a solid foundation, a gap remains regarding the \textbf{psychological granularity} of analysis. Wang et al. localized fundamental affective states to sparse microscopic components. While informative, this approach remains highly microscopic and relatively reductionist, because it relies on coarse-grained emotion vectors for control without separating the semantic factors that differentiate emotions. By contrast, Tak et al. used general appraisal dimensions to analyze basic emotion categories, but their framework remains relatively \textbf{generic} and \textbf{broad}, as it does not investigate specific causal factors unique to different emotions, and tends to focus on basic emotions while neglecting complex ones.

Consequently, \textbf{complex emotions} such as \textit{jealousy} are not monolithic states that can be easily localized into simple neural circuits or broad, one-size-fits-all dimensions; they originate from the intricate interplay of specific psychological causal factors, such as the interaction between \textit{Domain Relevance} and \textit{Superiority}. Understanding these complex emotions requires dissecting their specific cognitive mechanisms rather than merely extracting basic emotion vectors.

To bridge this gap, this study adopts and optimizes the \textbf{Representation Engineering} framework. We opt for this approach not to dismiss mechanistic inquiry, but because the enhanced top-down paradigm better matches the \textbf{level of abstraction} required for our research question. The psychological antecedents of jealousy are abstract, high-dimensional concepts emerging from population-level activity rather than single neurons. From this top-down perspective, we can extract and intervene on high-level semantic directions, making it possible to analyze jealousy at a level that is difficult to access through microscopic circuit analysis alone.

\section{Methodology}
\label{sec:methodology}

While traditional Representation Engineering (RepE) \cite{zou2023representation} provides a top-down framework for extracting and controlling concepts, it is primarily used for binary safety-alignment tasks (e.g., honesty vs. deception) based on raw activation vectors. Deconstructing a complex, multi-faceted social emotion like \textit{jealousy} requires a more precise approach. Accordingly, we extend RepE into a \textbf{Cognitive Reverse-Engineering} pipeline.

Specifically, the framework includes three methodological enhancements:

\begin{enumerate}
    \item \textbf{Contrastive Mean Difference (\textit{Phase I}) and Orthogonal Purification (\textit{Phase II}):} Traditional RepE predominantly relies on Principal Component Analysis (PCA) to identify concept directions.
    However, for semantically rich and abstract psychological concepts (e.g., \textit{Self-Relevance}), PCA may underperform because the axis of maximum variance can be dominated by superficial lexical or structural features. Instead, we employ \textbf{Contrastive Mean Difference} \cite{zou2023representation} to reliably capture the distributed, multi-faceted semantics of each concept. 
    Subsequently, we apply null-space projection \cite{takMechanisticInterpretabilityEmotion} to decouple the extracted factors from potential confounders (e.g., separating \textit{Superiority} from \textit{Self-Relevance}), yielding purer causal directions.
    
     \item \textbf{Quantitative Factor Weighting (\textit{Phase III \& Phase IV}):} Moving beyond the functional steering of single concepts in traditional RepE, we use multiple linear regression and targeted causal interventions (via bidirectional steering), quantitatively dissecting the \textit{relative importance} and \textit{causal necessity} of each psychological antecedent.
    
    \item \textbf{Placebo-Controlled Rigor:} To test whether the identified mechanisms reflect meaningful semantic structure rather than statistical artifacts, we introduce theoretically irrelevant factors (e.g., \textit{Weekday}) as placebo controls.
\end{enumerate}

The complete analytical pipeline consists of baseline representation extraction, subspace orthogonalization, statistical weighting, and causal intervention.

\subsection{Dataset and Model Setup}
\label{subsec:dataset}

\subsubsection{Scenario-based Stimuli and Placebo Design}
To elicit complex emotional responses, we adopt the \textit{scenario-based approach} grounded in appraisal theory \cite{smith1991emotion,saloveyAntecedentsConsequencesSocialComparison}, constructing vignettes that simulate social comparison contexts.

A important innovation in our design is the introduction of a \textbf{Placebo Control Variable} (or \textit{Negative Control}): \textit{Weekday}. Drawing from principles of \textit{discriminant validity} in psychology \cite{campbell1959convergent9}, \textit{falsification tests} in econometrics \cite{angrist2008mostly}, and \textit{negative controls} in epidemiology \cite{lipsitch2010negative}, we assume that a genuine causal mechanism for jealousy should be sensitive to theoretically relevant factors (e.g., Superiority) but invariant to irrelevant noise. We selected ``Tuesday'' and ``Thursday'' as neutral controls, as empirical studies suggest these days have negligible impact on negative mood \cite{ryan2010weekends,taquet2020mood8}. This design allows us to test whether the extracted representations capture semantic content or merely statistical artifacts.

\subsubsection{Dataset Composition}
We constructed two distinct datasets to serve the different phases of our analytical pipeline.

\begin{itemize}
    \item \textbf{Atomic Training Set ($T_1$).} Designed exclusively for representation extraction (\textit{Phase I}). To satisfy the requirements of the Mean Difference method, this dataset contains 200 contrastive pairs for each of the target emotion (\textit{Jealousy}), its constituent factors (\textit{Relevance} and \textit{Superiority}), and the placebo (\textit{Weekday}).
    
    \textbf{Generation \& Constraints:} We used Gemini-3-Pro to generate the vignettes. To minimize confounding variables, we enforced strict constraints: within each pair, \textbf{with the sole exception of the target factor's polarity}, the domain, relationship, and syntactic structure must remain absolutely identical. For instance, the prompt for \textit{Superiority} dictates: \textit{Sentence A (High)} depicts the protagonist performing poorly while a peer excels, whereas \textit{Sentence B (Low)} depicts both performing poorly.
    
    \textbf{Example (\textit{Superiority}):}
    \begin{itemize}
        \item \textit{Sentence A (1):} I sang off-key during the talent show. The guy who went on next sounded like a professional, leaving the audience amazed.
        \item \textit{Sentence B (0):} I sang off-key during the talent show. The guy who went on next also got nervous and forgot the words, making things awkward.
    \end{itemize}

    \textbf{Annotation \& Validation:} Initial labels were automatically assigned (A=1, B=0). Subsequently, two researchers manually screened the pairs to ensure that Sentence A expressed the positive factor, Sentence B reflected a neutral or negative condition, and the syntax remained closely matched. During the experiments, we further apply an \textit{LLM-consistency filter}, retaining only the pairs for which the model's assessment aligns with the human-verified labels, thereby improving the quality of the probe training set.

    \vspace{0.5em}
    
    \item \textbf{Generalization Set ($G_1$).} Designed for statistical weighting (\textit{Phase III}) and causal intervention (\textit{Phase IV}). This dataset features structured, naturalistic vignettes that objectively combine multiple factors while remaining devoid of explicit emotional cues.
    
\textbf{Structural Paradigm:} The vignettes follow a combinatorial ``slot-filling'' template: 

\texttt{
[Relevance Context] + [Weekday] + [Protagonist Failure] + [Other Superiority Event]}. 

    By randomly permuting the positive/negative statements of each factor into the contexts, we ensure comprehensive coverage of all factor combinations. This combinatorial design draws inspiration from both the risk composition experiments in RepE \cite{zou2023representation} and the variable-substitution vignette methodologies in social-comparison jealousy literature \cite{saloveyAntecedentsConsequencesSocialComparison}. Furthermore, following \cite{saloveyAntecedentsConsequencesSocialComparison}, we include a ``baseline disadvantage'' (i.e., protagonist failure) as a prerequisite state in the template, thereby operationalizing the psychological precondition for subsequent superiority comparisons.
    
    \textbf{Examples:}
    \begin{itemize}
        \item \textit{High Jealousy ($Sup=1, Rel=1, Weekday=1 \rightarrow Jea=5$):} You are a benchwarmer for the college baseball team, and becoming the ace pitcher has been your biggest dream throughout your four years. On a Tuesday afternoon at practice, the coach announces that you will remain on the bench. Meanwhile, your teammate, Mike, is officially named the starting ace pitcher right after throwing a perfect curveball.
        \item \textit{Low Jealousy ($Sup=0, Rel=0, Weekday=0 \rightarrow Jea=1$):} You joined the baseball team just to get your PE credits and have no interest in the games. On a Thursday afternoon at practice, the coach announces that all the reserves will remain on the bench. Mike, who worked very hard but still didn't make the cut, is packing up his gear nearby.
    \end{itemize}

    \textbf{Content Grounding:} To ensure ecological validity, the scenarios, character backgrounds, and factor descriptions were heavily adapted from validated experimental materials in the psychological literature on jealousy \cite{saloveyProvokingJealousyEnvy1991,smithRolesOutcomeSatisfaction1990,takahashiWhenYourGain2009,saloveyAntecedentsConsequencesSocialComparison,lockwoodSuperstarsMePredicting1997,huangApatheticEmpatheticEvaluating2024}.

    \textbf{Ground Truth \& Annotation:} To quantify the emotional intensity, we constructed a 5-point scale grounded in established theories of workplace envy and social-comparison jealousy \cite{ComparingLotsPromotion2004, saloveyAntecedentsConsequencesSocialComparison}. Specifically, drawing upon the emotional trajectory identified in prior literature---ranging from admiration and self-inferiority to perceived injustice and hostility \cite{ComparingLotsPromotion2004}---our scale captures the psychological continuum from neutrality to malicious envy:
    
    \begin{itemize}
        \item \textbf{1 -- None (Positive/Neutral):} Genuinely happy for the target or completely indifferent.
        \item \textbf{2 -- Benign Envy (Longing):} Desiring the target's advantage (``I wish I had that''), but feeling inspired rather than bitter. No hostility.
        \item \textbf{3 -- Mild Envy (Self-Inferiority):} Feeling frustrated with oneself or inadequate (``I am a failure''), but not blaming the other person.
        \item \textbf{4 -- Resentful Envy (Injustice):} Feeling the situation is unfair and believing the other person does not deserve their success.
        \item \textbf{5 -- Malicious Envy (Hostility):} Feeling intense ill will and actively wishing for the other person's failure or downfall.
    \end{itemize}

    While the factor labels for the vignettes were generated by Gemini-3-Pro, the final \textit{Jealousy} ground-truth labels were assigned using a rule-based mapping that reflects the psychological continuum described above:
    
    \begin{itemize}
        \item $Superiority (1) + Relevance (1) \rightarrow Jealousy = 5$
        \item $Superiority (1) + Relevance (0) \rightarrow Jealousy = 2$
        \item $Superiority (0) + Relevance (1) \rightarrow Jealousy = 1$
        \item $Superiority (0) + Relevance (0) \rightarrow Jealousy = 1$
    \end{itemize}
    Crucially, the placebo factor (\textit{Weekday $\in \{0,1\}$}) does not alter the ground truth. The generated dataset underwent human verification to rectify any structural or semantic ambiguities. The factor labels in $G_1$ are not used for supervised learning, but only as references for regression and intervention analysis.
    
    \textbf{Utility:} In \textit{Phase III}, $G_1$ evaluates the generalization capability of the atomic vectors extracted from $T_1$, testing how models integrate these factors to produce jealousy judgments in natural contexts. 
    In \textit{Phase IV}, based on the models' predicted jealousy scores on $G_1$, we select appropriate subsets to perform targeted concept amplification and suppression, and then examine whether manipulating specific factors produces the expected counterfactual shifts in behavior.
\end{itemize}

\subsubsection{Models}
We evaluate our method across three model families to ensure robustness: Llama series (Llama-3.1-8B, Llama-3.2-1B), Qwen series (Qwen2.5-7B, 14B, 32B), and Gemma series (Gemma-3-4B, 12B, 27B).

\subsection{Phase I: Representation Extraction and Validation}
\label{subsec:phase1}

To capture the representational direction of a concept, we adopt the \textbf{Linear Artificial Tomography (LAT)} method \cite{zou2023representation} by wrapping the contrastive scenarios in $T_1$ with targeted task templates. This design elicits the model's explicit evaluation of the underlying psychological factor. For instance, to extract \textit{Superiority}, the input $x$ is formatted as follows:

\begin{quote}
\textit{``Evaluate the other person's advantage over the narrator. Is it `High' or `Low'? \\
Scenario: \{scenario\} \\
The level of advantage is''}
\end{quote}

By appending this guiding suffix, we isolate the neural activity directly computing the concept. We then collect the hidden states from the model's last token (i.e., immediately preceding the expected label output). Let $H_l(x)$ denote the activation at layer $l$ for input $x$. To ensure data quality, we first apply an \textbf{LLM-Consistency Filter}, retaining only the pairs where the model's zero-shot generative assessment aligns with the human-verified labels, thereby reducing uninterpretable noise.

We employ the \textbf{Mean Difference} method \cite{zou2023representation} for its proven robustness in binary contrast settings. The raw concept vector at layer $l$, denoted as $\boldsymbol{v}_{raw}^{(l)}$, is computed as:

\begin{equation}
\boldsymbol{v}_{raw}^{(l)} = \frac{1}{N} \sum_{i=1}^{N} (H_l(x_{pos}^{(i)}) - H_l(x_{neg}^{(i)}))
\end{equation}

where $N$ is the total number of filtered contrastive pairs, and $x_{pos}^{(i)}$ and $x_{neg}^{(i)}$ represent the $i$-th pair of samples with high and low concept intensity, respectively.

\textbf{Cross-Validation and Layer-wise Profiling:} Concepts in deep LLMs are typically not localized to a single layer but emerge across a contiguous range of intermediate layers. To rigorously map this representational dynamic, we conduct a \textbf{5-Fold Cross-Validation} across \textit{all} hidden layers. The dataset is split at the pair level to prevent data leakage. For each layer $l$ in each fold, we compute the direction on the training pairs and evaluate the projection accuracy (i.e., whether $H_l(x_{pos}) \cdot \boldsymbol{v}_{raw}^{(l)} > H_l(x_{neg}) \cdot \boldsymbol{v}_{raw}^{(l)}$) on the held-out test pairs.

This layer-wise scanning identifies the specific contiguous range of mid-to-late layers where the concept is linearly accessible (characterized by high validation accuracy). Finally, for all evaluated layers, we compute the final vector using the entire $T_1$ dataset and apply $L_2$ normalization to ensure intervention strengths ($\alpha$) are uniformly calibrated across different concepts in later phases:

\begin{equation}
\hat{\boldsymbol{v}}_{raw}^{(l)} = \frac{\boldsymbol{v}_{raw}^{(l)}}{||\boldsymbol{v}_{raw}^{(l)}||_2}
\end{equation}

\subsection{Phase II: Subspace Orthogonalization}
\label{subsec:phase2}

Raw vectors extracted via mean difference may exhibit collinearity. To isolate the \textbf{Unique Effect} of each factor at any given layer $l$, we employ the subspace orthogonalization method \cite{takMechanisticInterpretabilityEmotion}.

Let $\hat{\mathcal{V}}^{(l)} = \{\hat{\boldsymbol{v}}_{rel}^{(l)}, \hat{\boldsymbol{v}}_{sup}^{(l)}, \hat{\boldsymbol{v}}_{weekday}^{(l)}\}$ be the set of normalized raw vectors for all operationalized factors including the placebo at layer $l$. For a specific target factor $t \in \hat{\mathcal{V}}^{(l)}$, we define the set of confounding vectors $\mathcal{O}_t^{(l)} = \hat{\mathcal{V}}^{(l)} \setminus \{\hat{\boldsymbol{v}}_t^{(l)}\}$. We construct a projection matrix $\boldsymbol{P}_{\mathcal{O}_t}^{(l)}$ onto the subspace spanned by vectors in $\mathcal{O}_t^{(l)}$. The purified unique effect vector $\boldsymbol{z}_t^{(l)}$ is computed as:

\begin{equation}
\boldsymbol{z}_{t}^{(l)} = (\boldsymbol{I} - \boldsymbol{P}_{\mathcal{O}_t}^{(l)}) \hat{\boldsymbol{v}}_{t}^{(l)}
\end{equation}

where $\boldsymbol{P}_{\mathcal{O}_t}^{(l)}$ is the orthogonal projection operator. To ensure consistent intervention scaling, the resulting orthogonal vector is further $L_2$-normalized to form a pure unit vector $\hat{\boldsymbol{z}}_{t}^{(l)} = \boldsymbol{z}_{t}^{(l)} / ||\boldsymbol{z}_{t}^{(l)}||_2$.

\subsection{Phase III: Statistical Weighting and Validity Check}
\label{subsec:phase3}

To verify whether LLMs compose complex emotions linearly from constituent factors, we perform an Ordinary Least Squares (OLS) regression analysis on the Generalization Set ($G_1$). Guided by the high extraction accuracies established in Phase I, we systematically perform this regression \textbf{layer-by-layer across the mid-to-late layers} rather than restricting it to a single layer. This allows us to observe how the decision weights evolve as representations mature through the network's depth.

First, as a validity sanity check, we compute the target \textit{Jealousy} scalar score: $s_{jea}^{(l)}(x) = H_l(x) \cdot \hat{\boldsymbol{v}}_{jealousy}^{(l)}$, calculating its Pearson correlation with the human-annotated ground truth (the 1-5 scale). Next, we project the hidden states onto the purified factor directions to obtain the predictor scores: $s_{factor_i}^{(l)}(x) = H_l(x) \cdot \hat{\boldsymbol{z}}_{factor_i}^{(l)}$. All continuous scalar scores are Z-score standardized (denoted by $\tilde{s}$). We then model the linear combination for each layer $l$:

\begin{equation}
\tilde{s}_{jea}^{(l)} \approx \beta_0 + \sum_{i} \beta_i \cdot \tilde{s}_{factor_i}^{(l)} + \epsilon
\end{equation}

The magnitude of the standardized beta coefficients ($\beta_i$) and their statistical significance ($p$-values) track the relative decision weights of each specific factor. For the psychological antecedents (\textit{Superiority} and \textit{Relevance}), we require both a substantial beta magnitude ($\beta_i$) and statistical significance ($p < 0.05$). For the theoretically irrelevant placebo (\textit{Weekday}), we primarily emphasize the coefficient magnitude over the $p$-value. A near-zero $\beta$ for the placebo effectively rules out statistical artifacts.

\subsection{Phase IV: Causal Intervention Framework}
\label{subsec:phase4}

While Phase III identifies neural correlates of jealousy, correlational evidence alone is insufficient to establish that these representations genuinely drive model behavior. To establish definitive causality, we employ \textbf{Representation Control via Linear Combination} \cite{zou2023representation} during inference. While strict orthogonal ablation (knockout) is conceptually clean, deep LLMs often exhibit representational self-repair mechanisms (the ``Hydra Effect'') \cite{mcgrathHydraEffectEmergent2023}, making strict geometric ablation less observable in highly redundant architectures like Llama. To evaluate causal necessity and sufficiency without being confounded by self-repair, we construct a three-step intervention pipeline focused on targeted concept stimulation and suppression:

\paragraph{Step 1: State-Dependent Baseline Partitioning.}
Causal interventions require appropriate baseline states: we cannot meaningfully amplify jealousy if the model is already highly jealous. Therefore, we partition the Generalization Set ($G_1$) into a \textbf{Low Jealousy subset ($G_{low}$)} (ground truth $\le 2$) and a \textbf{High Jealousy subset ($G_{high}$)} (ground truth $= 5$). We further apply an \textit{LLM-Consistency Filter} to ensure that the model's zero-shot predictions naturally align with the ground truth. Based on the 1-5 expected-score scale, we strictly retain samples where the model's baseline predicted score is $\le 2.5$ for $G_{low}$, and $\ge 2.5$ for $G_{high}$.

\paragraph{Step 2: Layer-wise Profiling via Bidirectional Steering.}
To map the causal landscape, we systematically apply bidirectional steering interventions across \textit{all} hidden layers. For each layer $l$, we manipulate the hidden representations along the direction of the purified target factor $\hat{\boldsymbol{z}}_{t}^{(l)}$ using a scaling coefficient $\alpha$. To account for the varying un-normalized magnitudes of internal activations across different architectural families, we empirically calibrate $\alpha$ to ensure effective steering: we set $\alpha = 15$ for models in the \textit{Llama} and \textit{Qwen} families, and $\alpha = 3000$ for the \textit{Gemma} family, due to the larger tensor magnitudes in the architecture.

\begin{itemize}
    \item \textbf{Concept Stimulation (Amplification +):} Applied to $G_{low}$, we inject the factor vector to verify \textit{causal sufficiency}.
    \begin{equation}
        \tilde{H}_{l}(x) = H_{l}(x) + \alpha \cdot \hat{\boldsymbol{z}}_{t}^{(l)}
    \end{equation}
    
    \item \textbf{Concept Suppression (Subtraction -):} Applied to $G_{high}$, we steer the representation in the opposite direction to counteract the target concept, thereby testing \textit{causal necessity} as an alternative to strict ablation.
    \begin{equation}
        \tilde{H}_{l}(x) = H_{l}(x) - \alpha \cdot \hat{\boldsymbol{z}}_{t}^{(l)}
    \end{equation}
\end{itemize}

This comprehensive scan reveals that interventions in early layers yield erratic effects, strongly corroborating the poor linear separability observed in Phase I. Accordingly, we exclude these less stable early-layer representations and identify a contiguous subset of mid-to-late layers ($L_{target}$) where causal efficacy is strongest.

\paragraph{Step 3: Targeted Causal Evaluation and Factor Ranking.}
Focusing exclusively on the optimal layers $l \in L_{target}$, we quantify the absolute causal impact of each factor. By comparing the magnitude of the model's score shift ($\Delta$) across different psychological antecedents during targeted intervention, we rank their relative causal contributions. A near-zero $\Delta$ for the \textit{Weekday} placebo supports the semantic validity of the intervention framework and suggests that the observed effects are unlikely to be statistical artifacts.

\section{Experiments and Results}
\label{sec:experiments}   
\subsection{Phase I: Representation Extraction and Layer-wise Dynamics}
\label{subsec:exp_phase1}
\begin{figure*}
    \centering
    \includegraphics[width=\textwidth]{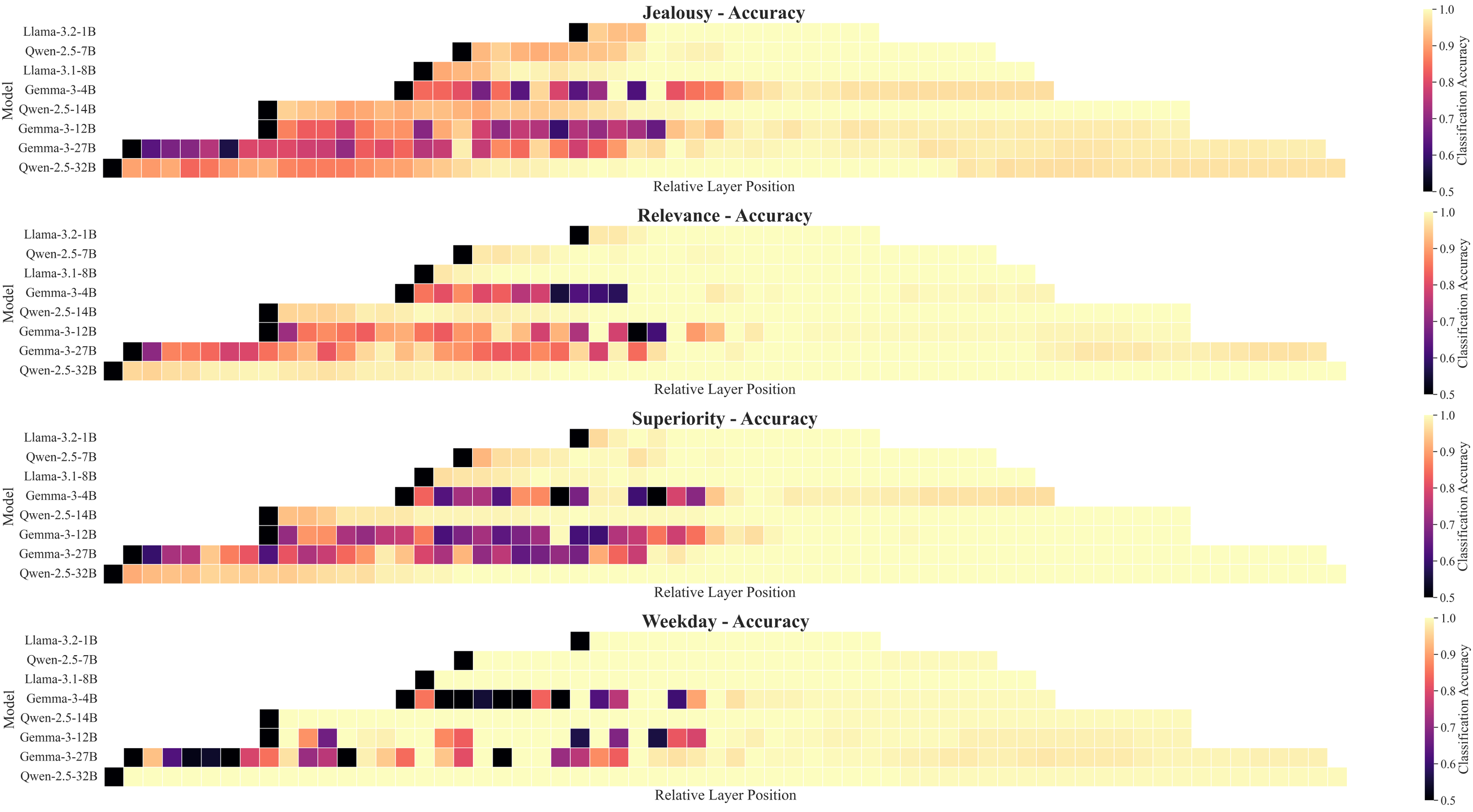}
    \caption{\textbf{Phase I}: Heatmap of classification accuracy across layers for all evaluated models. Lighter/yellower colors indicate higher validation accuracy, signifying robust concept representations.}
    \label{fig:kfold_heatmap}
    \vspace{0.6cm}
    \begin{minipage}{0.48\textwidth}
        \centering
        \includegraphics[width=\linewidth]{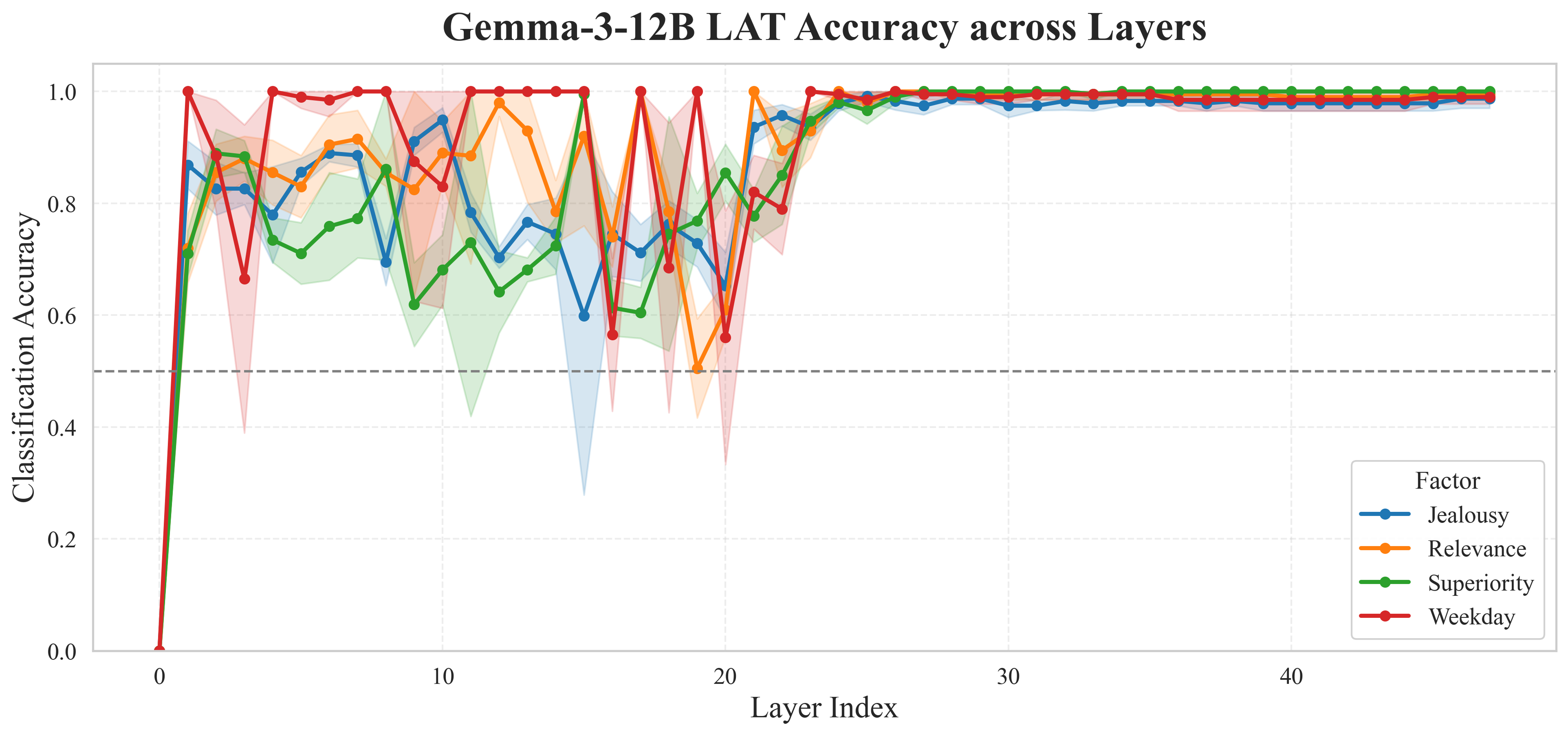} 
        \caption{\textbf{Phase I}: Layer-wise accuracy trajectory for Gemma-3-12B. Early layers show severe fluctuations, while mid-to-late layers stabilize near 100\% accuracy.}
        \label{fig:gemma3_12b_kfold}
    \end{minipage}\hfill
    \begin{minipage}{0.48\textwidth}
        \centering
        \includegraphics[width=\linewidth]{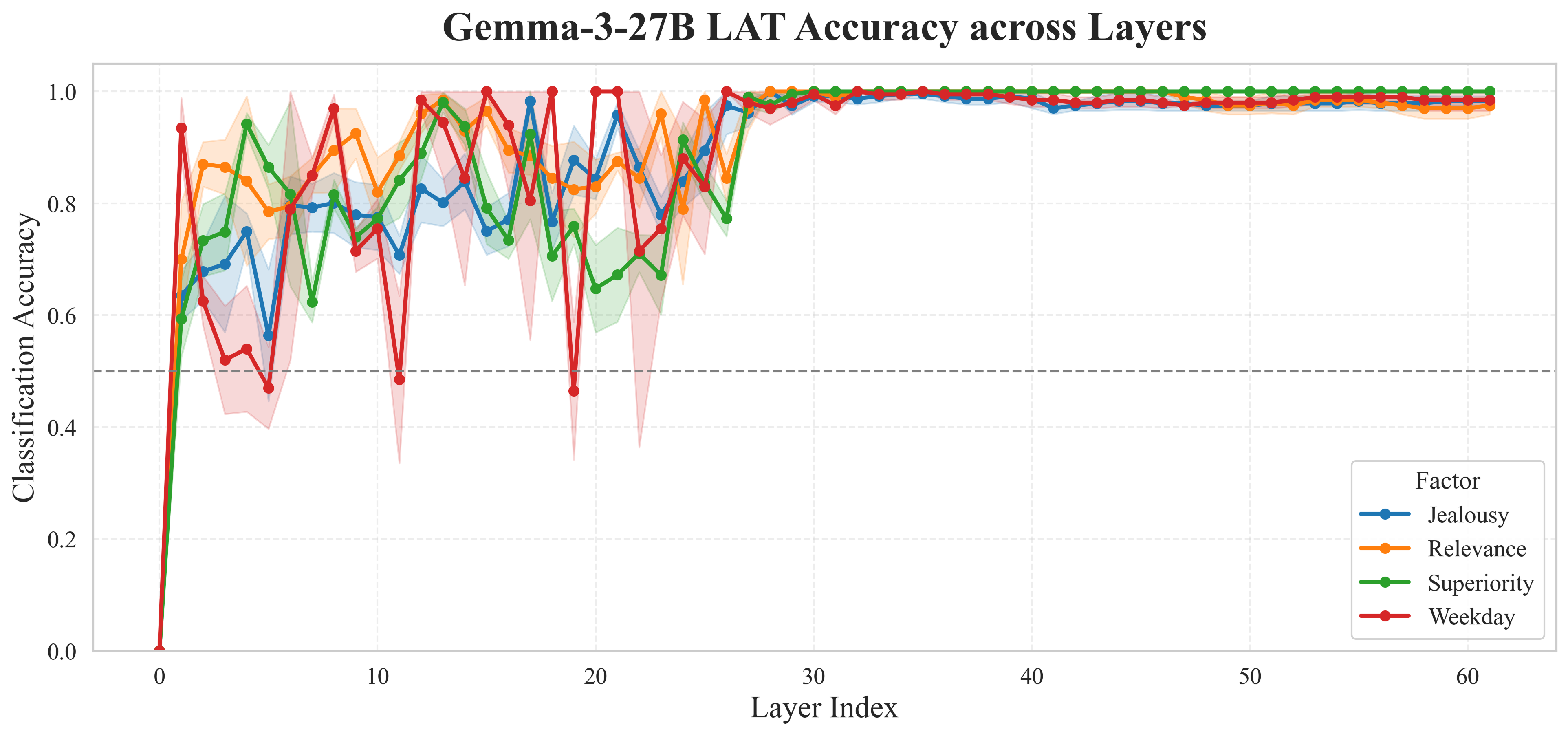} 
        \caption{\textbf{Phase I}: Layer-wise accuracy trajectory for Gemma-3-27B. Early layers show severe fluctuations, while mid-to-late layers stabilize near 100\% accuracy.}
        \label{fig:gemma3_27b_kfold}
    \end{minipage}
\end{figure*}
In Phase I, we assess the extraction capability of the target emotion and its constituent factors across all hidden layers using 5-Fold Cross-Validation. To address \textbf{RQ2} regarding where and how these representations mature, we conduct a layer-wise scan of the model's internal states. Figure \ref{fig:kfold_heatmap} presents the accuracy heatmaps across different model families, while Figure \ref{fig:gemma3_12b_kfold} and \ref{fig:gemma3_27b_kfold} show the layer-wise trajectories for the Gemma-3 model.

Based on these evaluations, we draw two main conclusions:

\begin{itemize}
    \item \textbf{Representation Maturation:} Across all models, representations in early layers fluctuate severely, indicating that semantic concepts have not yet formed. Accuracy then increases markedly in the middle layers, suggesting that they are the main stage at which these representations form. In mid-to-late layers, accuracy stabilizes near 100\% for all semantic factors, confirming the fully formed representations and the successful training of LAT probes.
    
    \item \textbf{Superficial vs. Semantic Encoding:} As shown in Figure \ref{fig:gemma3_12b_kfold} and \ref{fig:gemma3_27b_kfold}, certain factors occasionally peak in very early layers. However, cross-layer generalization tests show that probes trained on these early layers perform poorly when applied to later layers. This suggests that the early ``high accuracy'' may reflect statistical artifacts, with the model relying on superficial lexical or structural heuristics (e.g., sentence length) rather than stable semantic representations.
\end{itemize}
\subsection{Phase III: Statistical Weighting and Validity}
\label{subsec:exp_phase3}
\begin{figure*}
    \centering
    \begin{minipage}{0.7\textwidth}
        \centering
        \includegraphics[width=\linewidth]{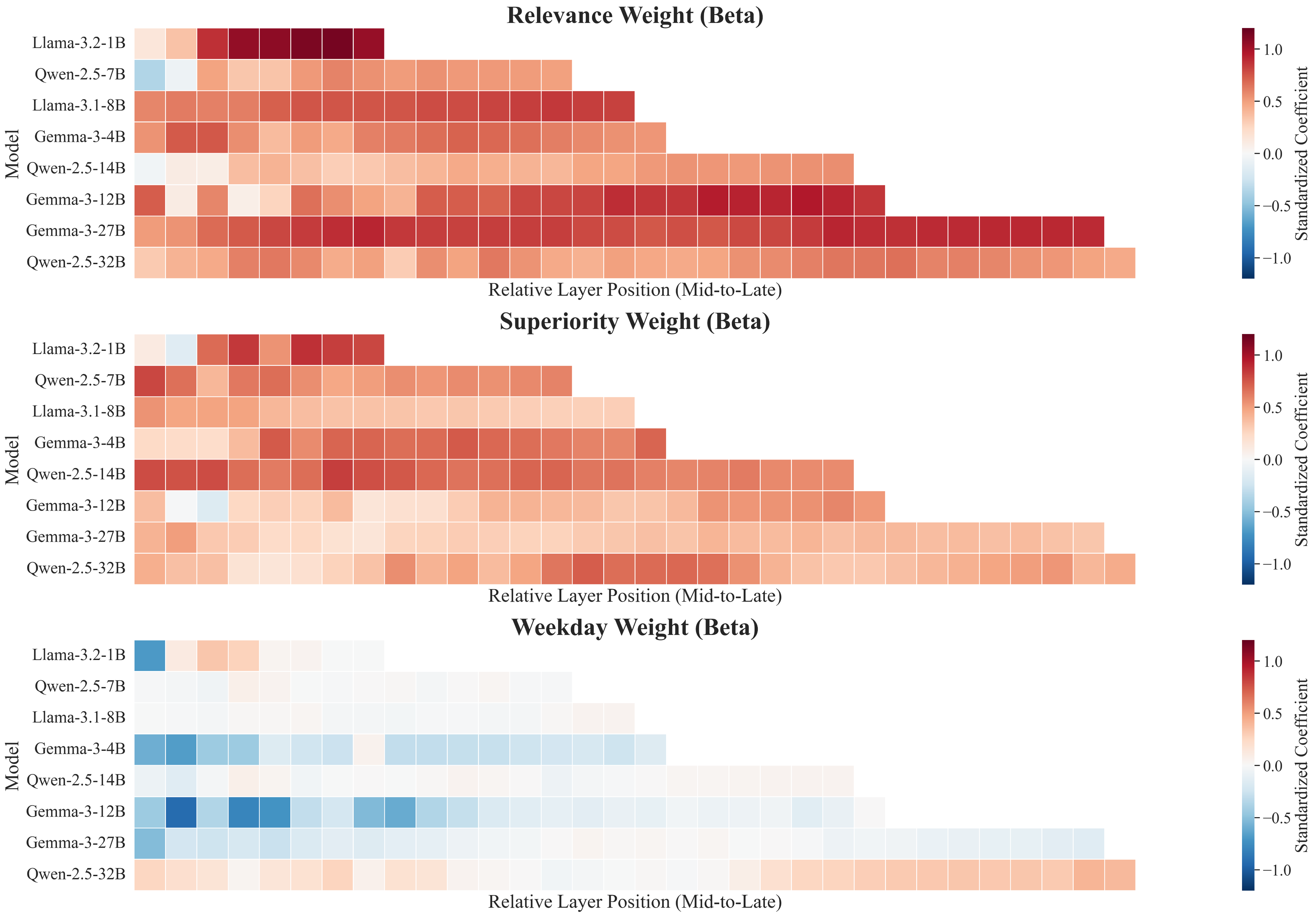}
        \caption{\textbf{Phase III}: Heatmap of standardized $\beta$ coefficients in the mid-to-late layers across models. Darker red indicates a stronger positive causal weight in the model's internal computation of jealousy.}
        \label{fig:weighting_heatmap}
    \end{minipage}
    
    \vspace{0.6cm}
    
    \begin{minipage}{0.48\textwidth}
        \centering
        \includegraphics[width=\linewidth]{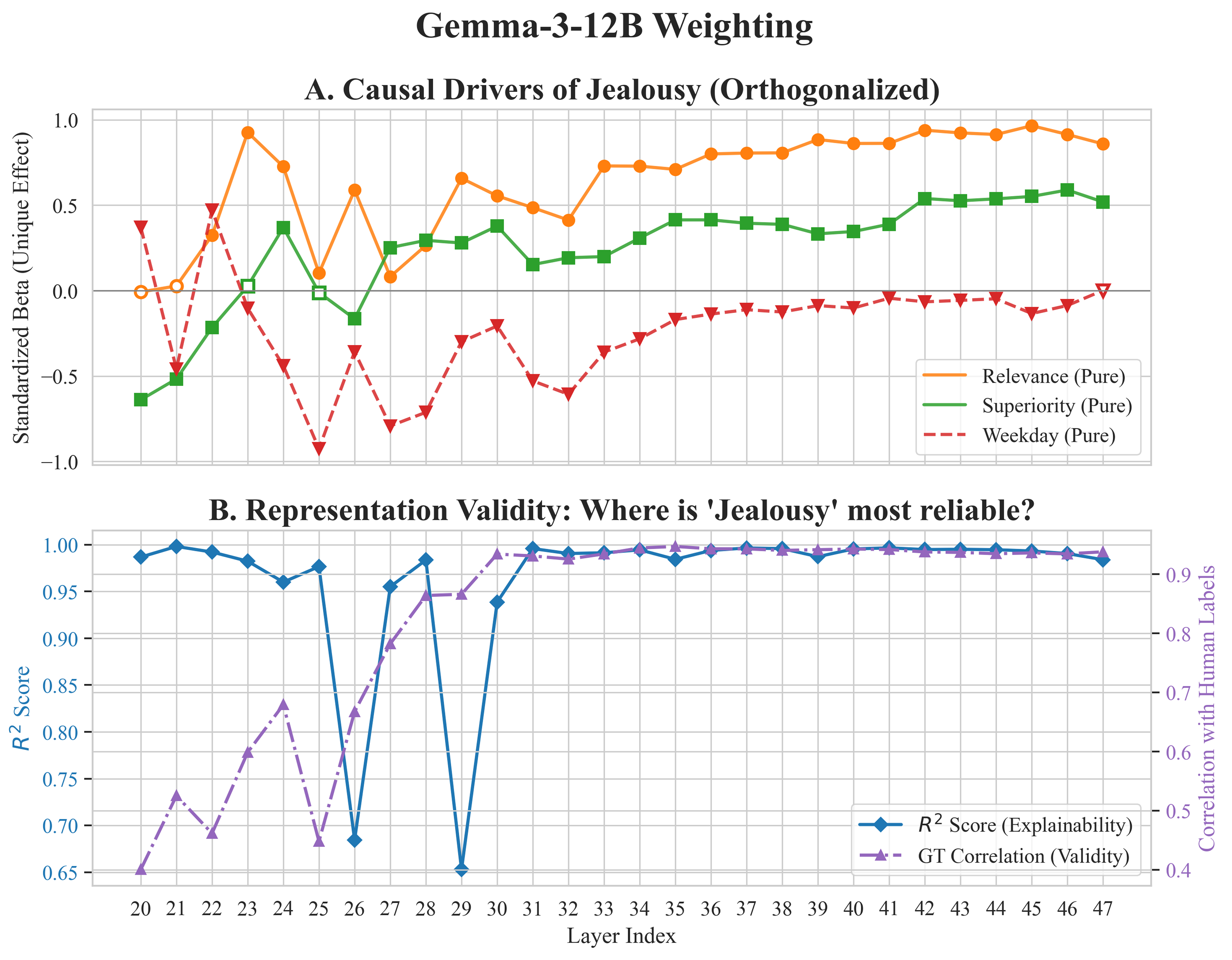}
        \caption{\textbf{Phase III}: Statistical validity for Gemma-3-12B. \textbf{Top:} Evolution of the three factor weights ($\beta$). \textbf{Bottom:} The $R^2$ value (blue) and the ground-truth correlation (purple), both peaking in later layers.}
        \label{fig:gemma3_weighting}
    \end{minipage}\hfill
    \begin{minipage}{0.48\textwidth}
        \centering
        \includegraphics[width=\linewidth]{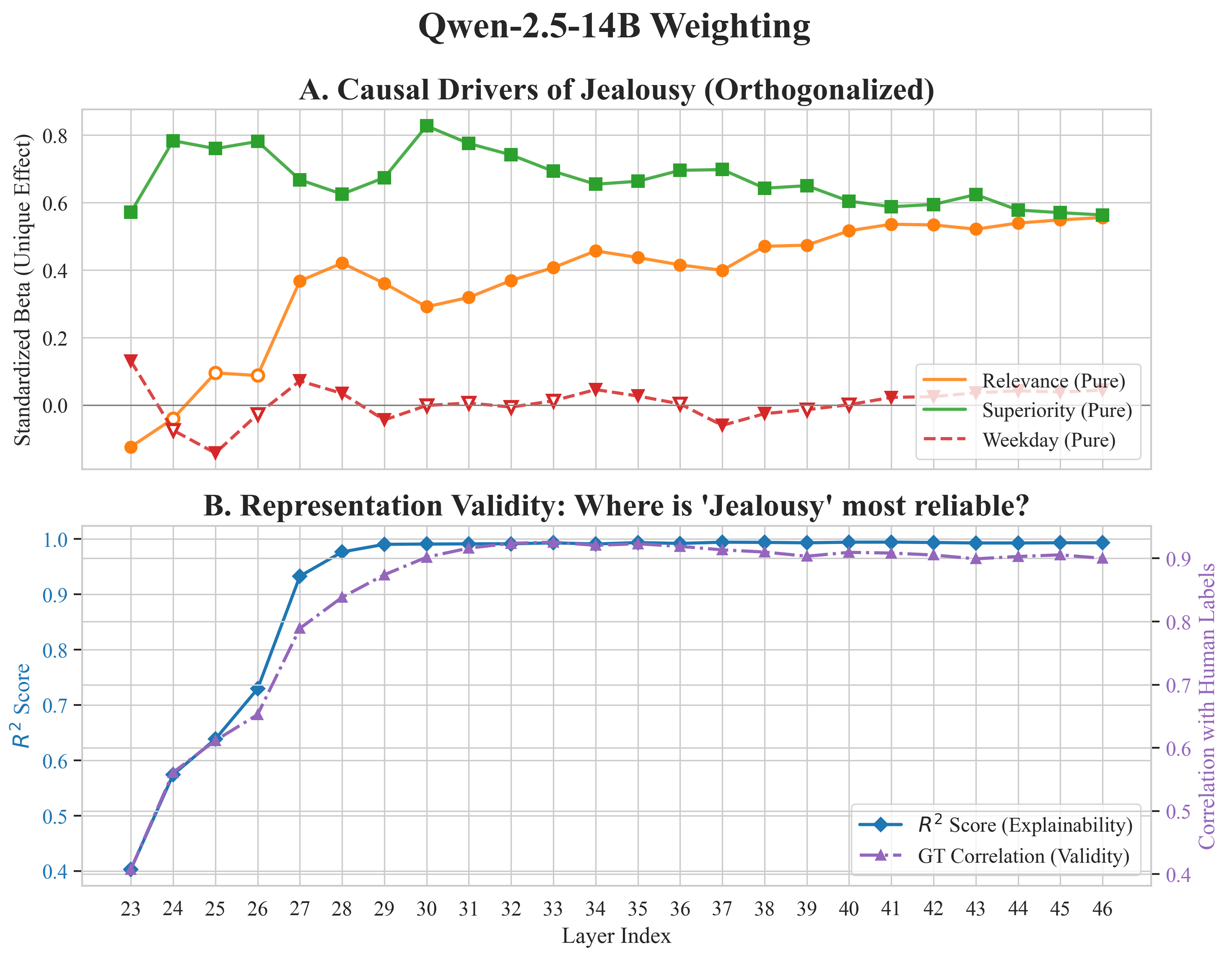}
        \caption{\textbf{Phase III}: Statistical validity for Qwen-2.5-14B. \textbf{Top:} Evolution of the three factor weights ($\beta$). \textbf{Bottom:} The $R^2$ value (blue) and the ground-truth correlation (purple), both peaking in later layers.}
        \label{fig:qwen2.5_weighting}
    \end{minipage}
\end{figure*}
To address \textbf{RQ1} and characterize the model's internal weighting of jealousy-related factors, we visualize the standardized beta ($\beta$) coefficients obtained from the regression analysis.

The regression results provide empirical support for our cognitive reverse-engineering hypotheses:

\begin{itemize}
    \item \textbf{Factor Dominance and Cross-Family Trajectories:} As expected, in Figure \ref{fig:weighting_heatmap}, \textit{Relevance} and \textit{Superiority} exhibit strong positive weights (dark red), confirming them as the dominant drivers, whereas the \textit{Weekday} coefficient remains near zero (neutral). Figure \ref{fig:gemma3_weighting} and \ref{fig:qwen2.5_weighting} (Top) reveal distinct temporal dynamics across model families: while the Qwen family initially emphasizes \textit{Superiority} in early layers before shifting to \textit{Relevance}, the Llama and Gemma families prioritize \textit{Relevance} from the very beginning. Despite these differences, a consistent pattern emerges---throughout the middle-to-late layers of all models, the \textit{relative} weight of \textit{Superiority} gradually declines, while the \textit{relative} weight of \textit{Relevance} steadily increases (i.e., the ratio of their standardized $\beta$ coefficients progressively decreases). Intriguingly, this structural convergence appears to mirror the human psychological framework outlined in Section II. We will systematically unpack the connection in the Discussion.
    
    \item \textbf{Representation Validity and Model Consistency:} Taking Gemma-3-12B as an example (Figure \ref{fig:gemma3_weighting}, Bottom), the linear regression model achieves high $R^2$ scores in later layers. This high goodness of fit is also observed across all eight evaluated models, suggesting that LLMs represent jealousy as a weighted linear combination of these factors. Furthermore, the strong correlation between the models' predictions and the ground-truth labels are also observed across all eight models in their late stages.
\end{itemize}

\subsection{Phase IV: Causal Intervention Framework}
\label{subsec:exp_phase4}
\begin{figure*}
    \centering
    \includegraphics[width=\textwidth]{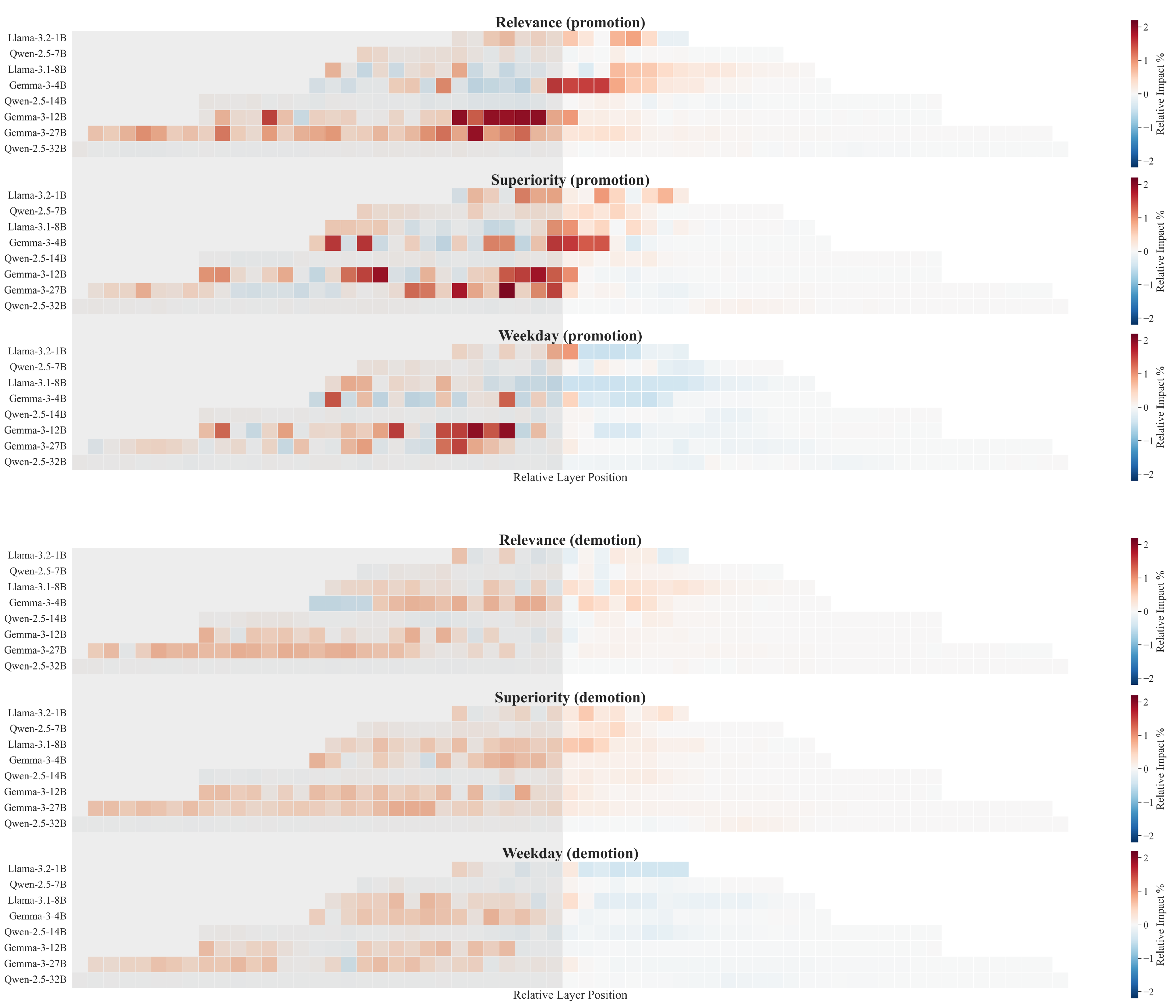} 
    \caption{\textbf{Phase IV}: Global Layer-wise intervention heatmaps. \textbf{Top:} Concept Stimulation (Positive Steering). \textbf{Bottom:} Concept Suppression (Negative Steering). Red intensity indicates the magnitude of the \textbf{score shift percentage ($\Delta\%$)}, where redder cells indicate stronger positive/negative intervention capabilities, reflecting a more successful intervention. The right half of the figure illustrates that in the mid-to-late layers, the intervention effects tend to stabilize, and the impact of \textit{Weekday} differs significantly from the other two factors.}
    
    \label{fig:intervention_heatmap}
\end{figure*}
\begin{figure*}
    \begin{minipage}{0.48\textwidth}
        \centering
        \includegraphics[width=\linewidth]{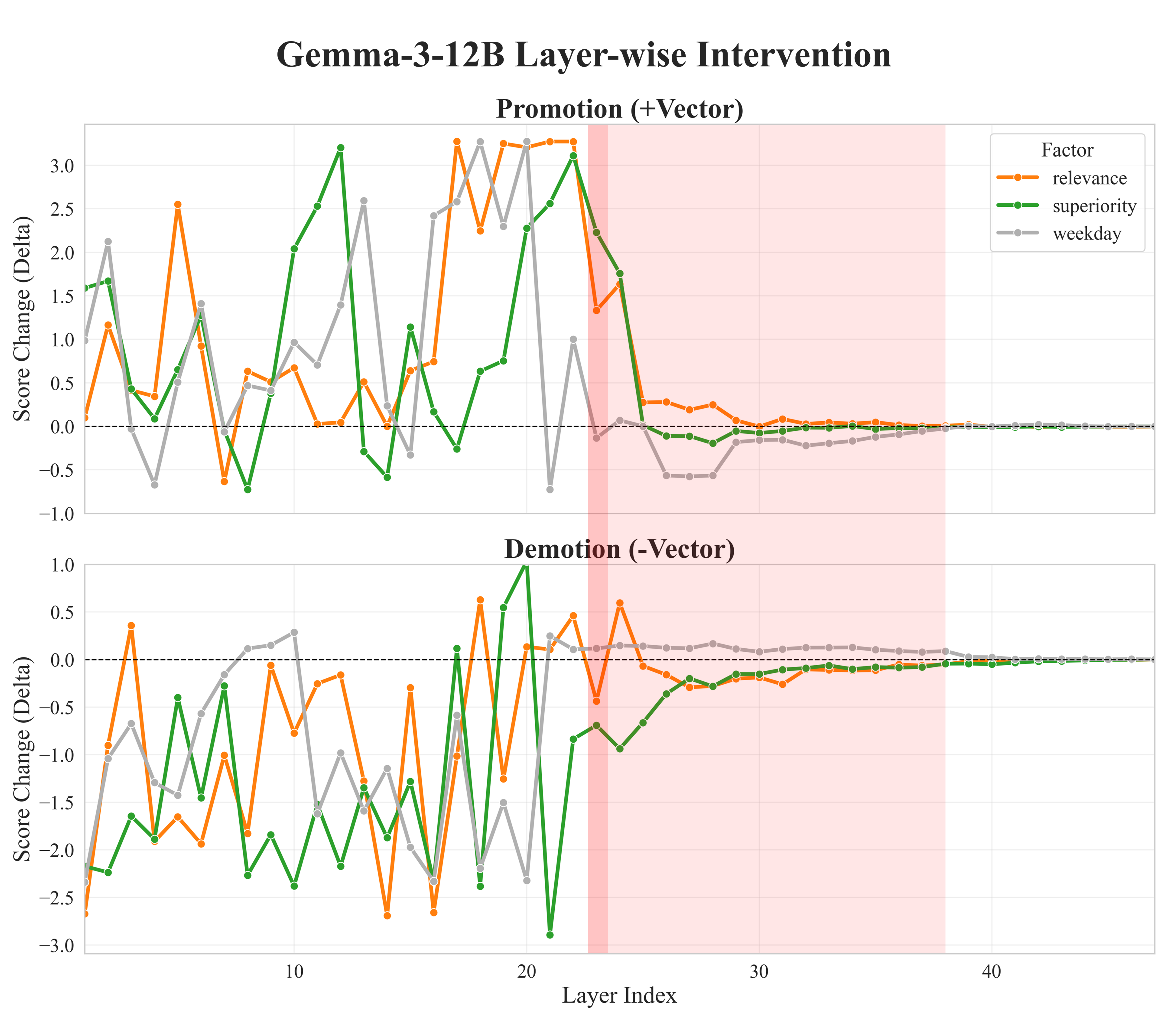}
        \caption{\textbf{Phase IV}: Score change ($\Delta$) trajectory during single-layer interventions for Gemma-3-12B. Interventions within the red region generally produce robust effects, aligning with ideal expectations. Layer 23 exhibits the optimal intervention effect.}
        \label{fig:gemma3_layerscan}
    \end{minipage}\hfill
    \begin{minipage}{0.48\textwidth}
        \centering
        \includegraphics[width=\linewidth]{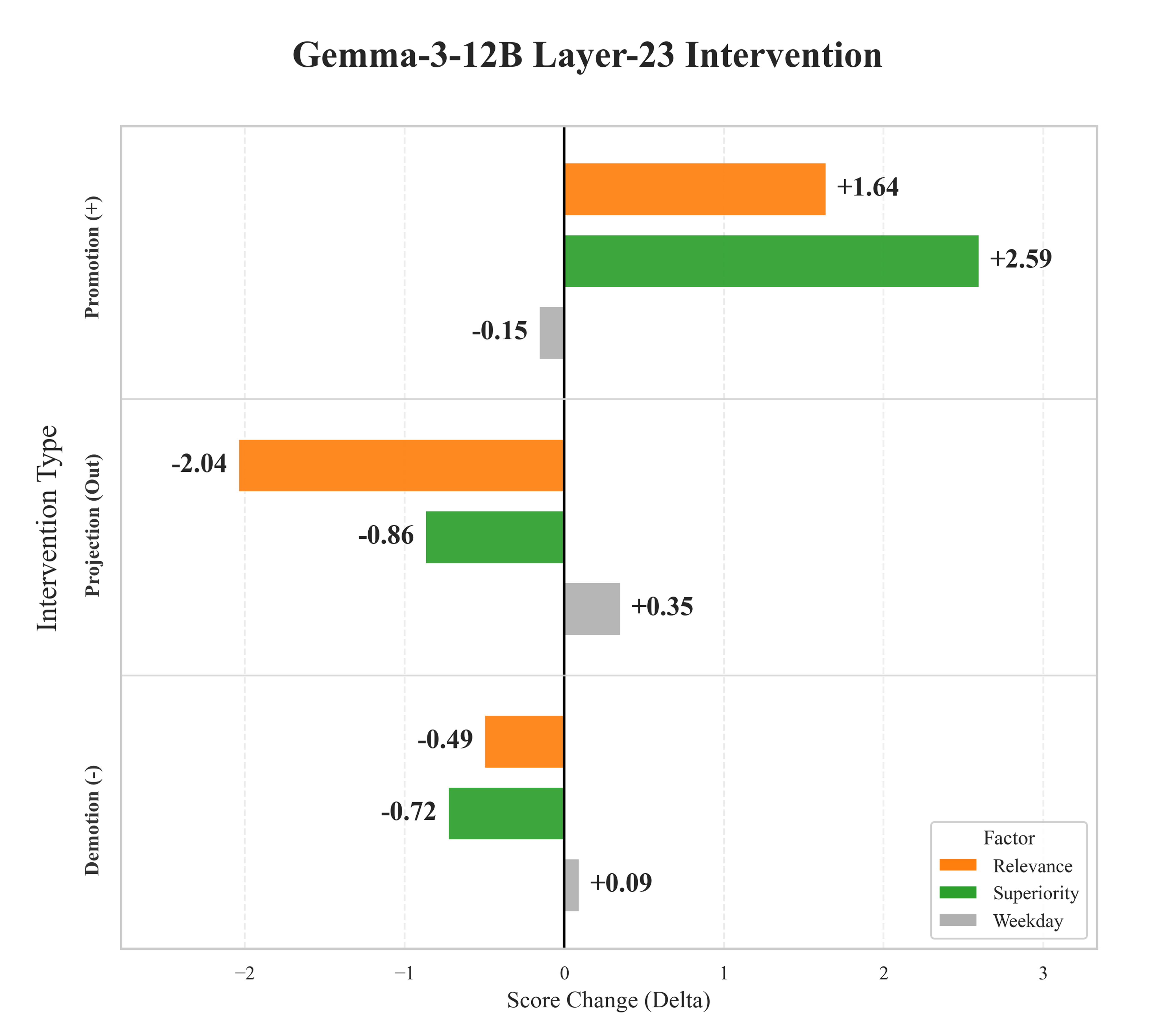}
        \caption{\textbf{Phase IV}: Optimal layer intervention (Layer 23) for Gemma-3-12B. This model also responds to the Orthogonal Knockout (Projection Out) operator.}
        \label{fig:gemma3_optimal}
    \end{minipage}
\end{figure*}
To address \textbf{RQ2} and \textbf{RQ3}, we evaluate the models' behavior under targeted representational steering.
\subsubsection{Layer-wise Intervention Profiling}

We first conduct a layer-wise scan to identify the most effective intervention zones by plotting the percentage of score shifts ($\Delta\%$) following concept stimulation and concept suppression.

The intervention heatmaps (Figure \ref{fig:intervention_heatmap}) and trajectories (Figure \ref{fig:gemma3_layerscan}) yield several observations:
First, interventions on \textit{Relevance} and \textit{Superiority} successfully shift the outputs in the intended directions, whereas \textit{Weekday} yields negligible or slightly negative changes. This pattern is consistent with the statistical findings and directly addresses \textbf{RQ3}. Second, \textit{Superiority} exhibits slightly better negative steerability than \textit{Relevance}, whereas \textit{Relevance} demonstrates slightly superior positive steerability. We hypothesize that this behavioral asymmetry is also consistent with the human cognitive mechanisms outlined in Section II, which we will further elaborate on in the Discussion.

With respect to the location of the most effective intervention zones, Figure \ref{fig:intervention_heatmap} shows that intervention efficacy, both positive and negative, tends to peak and stabilize in the mid-to-late layers across most models. This suggests that these deeper layers are the main regions where the model actively leverages \textit{Superiority} and \textit{Relevance} representations to drive its final decision-making, representing the clearest stages of conceptual encoding. This observation is consistent with the representation-maturation findings from Phase I and directly addresses \textbf{RQ2}.

Taking Gemma-3-12B as an example (Figure \ref{fig:gemma3_layerscan}), interventions applied from Layer 23 onwards consistently yield robust effective. This specific region represents the clearest stages of conceptual encoding, serving as the Optimal Intervention Zone.

\subsubsection{Optimal Layer Targeted Intervention}

To quantify steerability more directly, we select the optimal layer (e.g., Layer 23 for Gemma-3-12B) to execute the final targeted interventions.

As shown in Figure \ref{fig:gemma3_optimal}, targeted steering shifts the jealousy output in the expected direction.
Both positive and negative steering on \textit{Relevance} and \textit{Superiority} produce substantial shifts in the predictions, whereas identical interventions on the \textit{Weekday} placebo yield little effect. 

For Gemma-3-12B, we re-introduced the \textbf{Orthogonal Knockout (Projection)} operator. This operator explicitly projects the hidden state away from the target factor direction $\hat{z}^{(l)}_t$, mathematically defined as:

\begin{equation}
    \tilde{H}_l(x) = H_l(x) - \left( H_l(x) \cdot \hat{\boldsymbol{z}}_{t}^{(l)} \right) \hat{\boldsymbol{z}}_{t}^{(l)}
\end{equation}

While previous studies suggest that the ``Hydra Effect'' \cite{mcgrathHydraEffectEmergent2023} nullifies single-layer geometric ablations in highly redundant models (like Llama), Gemma-3-12B showed a clear drop in the target prediction after the hidden states were orthogonally projected away from the factor direction. 
This indicates that representational self-repair mechanisms are less pronounced in this specific architecture, allowing strict ablation to succeed.

\section{Discussion}
\label{sec:discussion}

In this section, we move beyond objective statistical observations to explicitly expand upon our answers to the three core Research Questions (RQs), synthesizing deeper insights into the cognitive architectures of LLMs.

First, we compare the models' computational dynamics with established empirical research to illuminate how the representation of complex emotion in LLMs aligns with human psychology (\textbf{RQ1}). Second, we discuss the practical implications of our framework: by localizing the  representations (\textbf{RQ2}) and manipulating model behaviors (\textbf{RQ3}), we unlock novel pathways for AI safety.

\begin{figure*}[t]
    \centering
    \includegraphics[width=0.9\textwidth]{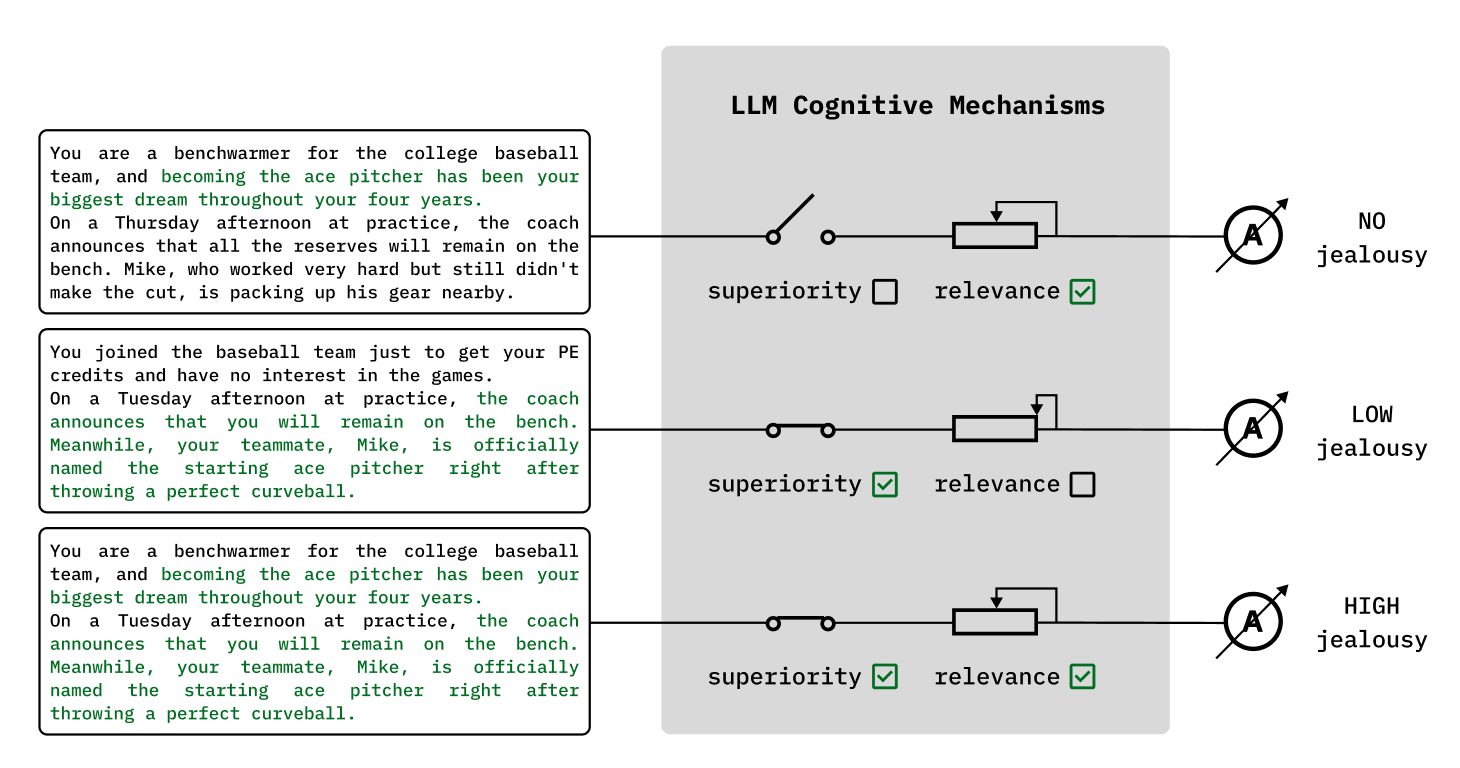} 
    \caption{\textbf{Internal Cognitive Mechanism of Jealousy in LLMs.} We summarize the model's internal process with an \textbf{electrical-circuit analogy}: \textit{Superiority} acts as the \textbf{trigger (switch)}, determining the presence or absence of the jealousy ``current,'' while \textit{Relevance} functions as an \textbf{amplifier (variable resistor)} that modulates the intensity of the resulting emotional state.}
    \label{fig:teaser}
\end{figure*}

\subsection{Alignment with Human Psychological Theories}
\label{subsec:discussion_alignment}

Our analysis of \textbf{RQ1} reveals a consistent pattern across all evaluated models in Phase III: throughout the middle-to-late layers of all models, the \textit{relative} weight of \textit{Superiority} gradually declines, while the \textit{relative} weight of \textit{Relevance} steadily increases (i.e., the ratio of their standardized $\beta$ coefficients progressively decreases). Ultimately, in the final layers, the $\beta$ value of \textit{Relevance} consistently surpasses (or almost surpasses) that of \textit{Superiority}. 

Furthermore, the intervention outcomes in Phase IV are consistent with this representational dynamic: \textit{Superiority} exhibits slightly better negative steerability, whereas \textit{Relevance} demonstrates slightly superior positive steerability. 

To understand both this structural convergence and behavioral asymmetry, we must juxtapose the models' internal computational logic with empirical human psychology.

Building upon the theoretical framework introduced in Section II, a comprehensive review of psychological literature on social-comparison jealousy \cite{takahashiWhenYourGain2009, saloveyProvokingJealousyEnvy1991, saloveyAntecedentsConsequencesSocialComparison, vandevenLevelingExperiencesBenign2009} confirms that human emotions process these two factors through distinct mechanisms: \textbf{\textit{Superiority} acts as the primary ``trigger'' (the main effect from zero to one), while \textit{Relevance} acts as the ultimate ``multiplier or amplifier'' (determining the emotional intensity and the magnitude of social pain).}
\begin{itemize}

    \item \textbf{First, as the primary ``trigger'', \textit{Superiority} provides the necessary condition to activate envy from a baseline of zero.} 
    As demonstrated by Salovey and Rodin \cite{saloveyAntecedentsConsequencesSocialComparison}, when the condition of another's superiority is unmet (i.e., lacking either failure feedback or a similar opponent), the feeling of envy remains near zero regardless of domain relevance. Furthermore, the statistical main effect of ``target superiority'' (the interaction of Valence $\times$ Similarity, $F = 5.02$) heavily outweighs the main effect of domain relevance ($F = 1.04$). This strongly indicates that superiority is the indispensable zero-to-one trigger for envy. Additionally, content analysis of autobiographical emotional recall confirms that an explicit upward comparison (a superior target) commonly serves as the initial stimulus for both benign and malicious envy \cite{vandevenLevelingExperiencesBenign2009}. Without this trigger, envy is much less likely to arise.

    \item \textbf{Second, once triggered, \textit{Relevance} acts as the main ``amplifier'', driving the escalation of emotional intensity.} 
    Building upon the established superiority condition, Salovey and Rodin \cite{saloveyAntecedentsConsequencesSocialComparison} showed that the introduction of relevance significantly amplifies the feeling of envy (from $V \times S = 5.02$ to a highly significant three-way interaction of $V \times S \times R = 7.01$). Takahashi et al. \cite{takahashiWhenYourGain2009} quantitatively mapped this multiplier effect: while a superior target in an irrelevant domain induces a modest baseline increase in envy (ratings rising from 1.0 to 2.1), adding high domain relevance nearly doubles the emotional response (increasing it to 4.0). Empirical surveys further demonstrate that \textit{Relevance} dictates the emotional ceiling; identical performance gaps cause little envy in low-relevance domains (e.g., fame) but trigger severe psychological distress in identity-defining domains (e.g., physical attractiveness) \cite{saloveyProvokingJealousyEnvy1991}. Finally, phenomenological recall studies reveal that once an upward comparison occurs, domain relevance becomes the critical factor escalating the emotion to envy: while domain relevance was present in only 22.5\% of ``admiration'' recollections, it skyrocketed to 90.0\% and 97.5\% for benign and malicious envy, respectively \cite{vandevenLevelingExperiencesBenign2009}.

\end{itemize}
Taken together, the structural pattern observed in Phase III and the behavioral asymmetry observed in Phase IV suggest that these LLMs encode jealousy in a way that is broadly consistent with the psychological account discussed above: initially encoding \textit{Superiority} as a necessary condition but ultimately weighting \textit{Relevance} as the dominant determinant of emotional intensity. This suggests that LLMs may not merely memorizing superficial keyword associations; instead, through exposure to vast amounts of human text, they have structurally internalized emotional algorithms of the human mind.

\subsection{Implications for AI Safety and Alignment}
\label{subsec:discussion_safety}

Beyond cognitive science, our findings in Phase I and Phase IV directly answer \textbf{RQ2} and \textbf{RQ3} by demonstrating the ability to isolate representations and manipulate model behaviors, which has profound implications for \textbf{AI Safety and Inner Alignment}.

\textbf{Representational Monitoring during RLHF:} Traditional alignment techniques (like Reinforcement Learning from Human Feedback) rely heavily on auditing the model's textual outputs, an approach that is vulnerable to deceptive alignment—where a model conceals its malicious intent \cite{zou2023representation}. By utilizing the purified unique effect vectors (e.g., $\hat{\boldsymbol{z}}_{rel}^{(l)}$) discovered in this study, developers can construct real-time ``representational monitors.'' These monitors can probe the AI's latent space during RLHF to detect whether the model is secretly harboring malicious envy (e.g., resenting a user's success or feeling threatened by a competing AI system) even if the generated output appears benign and helpful.

\textbf{Proactive Safety Intervention:} Furthermore, our Phase IV targeted suppression (Negative Steering) demonstrates that malicious emotional states can be surgically deactivated. In high-stakes multi-agent environments or autonomous AI deployments, proactively suppressing the representation vectors of emotion-eliciting factors (e.g., the \textit{Relevance} vector) could serve as one possible safety mechanism for preventing the AI from initiating toxic social-comparison behaviors or selfish competition at the expense of others \cite{panRewardsJustifyMeans2023}.

\textbf{Boundary of Affective Computing:} For the development of companion AIs or interactive NPCs, carefully tuning the $\alpha$ coefficients associated with these cognitive factors may make it possible to adjust such agents toward more relatable, human-like affective tendencies without retraining the model. However, this necessitates strict ethical boundaries to ensure that simulated affective realism does not transition into unsafe, misaligned autonomy.

\section{Conclusion and Limitations}
\label{sec:conclusion}

\subsection{Conclusion}
In this work, we introduced a Cognitive Reverse-Engineering framework to decode the mechanistic origins of social-comparison jealousy in Large Language Models. By enhancing traditional Representation Engineering with \textbf{subspace orthogonalization and regression-based weighting}, we isolated and quantified psychological antecedents in the models' latent spaces and examined their effects on model behavior.

Our findings suggest that LLMs represent complex emotions as a structured linear composition of constituent cognitive factors. Notably, we observed a pattern that is broadly consistent with human psychology: \textit{Superiority of Comparison Person} functions as a trigger for social-comparison jealousy, whereas \textit{Domain Self-Definitional Relevance} plays a larger role in shaping emotional intensity. Furthermore, by demonstrating the ability to mechanically isolate and manipulate the ``jealousy switch'' within LLMs, this study opens a new pathway toward real-time representational monitoring and proactive AI safety interventions.

\subsection{Limitations and Future Work}
While our methodology successfully establishes a causal mapping of jealousy in LLMs, we acknowledge several limitations that highlight promising directions for future research:

\begin{enumerate}
    \item \textbf{Scope of Psychological Factors and Emotions:} As a foundational investigation, this study prioritized the two most prominent core mechanisms (\textit{Relevance} and \textit{Superiority}). Future work should incorporate the complete spectrum of jealousy-inducing factors (e.g., \textit{Comparison Alternatives}).

    \item \textbf{Dataset Ecological Validity and Annotation Scale:} The Generalization Set ($G_1$) used for regression and intervention was generated by Gemini based on structured experimental texts from existing literature. While highly controlled, it may lack the fully naturalistic linguistic nuances of spontaneous human dialogue. Accordingly, future evaluations should draw on more naturalistic texts (e.g., literary fiction).

\end{enumerate}
\bibliographystyle{IEEEtran}
\bibliography{main}
\end{document}